\documentclass[sigconf]{acmart}

\usepackage[utf8]{inputenc}
\usepackage{multirow}   % 跨行支持
\usepackage{xcolor}     % 颜色支持
\usepackage{colortbl}   % 单元格背景色

\definecolor{highlightblue}{rgb}{0.9, 0.98, 1}
\definecolor{highlightgreen}{rgb}{0.9, 0.98, 0.9}
\AtBeginDocument{%
  }

\setcopyright{cc}
\setcctype{by}
\copyrightyear{2026}
\acmYear{2026}
\acmDOI{10.1145/3767308.3835719}
\acmConference[MM '26]{Proceedings of the 34th ACM International Conference on Multimedia}{November 10--14, 2026}{Rio de Janeiro, Brazil}
\acmBooktitle{Proceedings of the 34th ACM International Conference on Multimedia (MM '26), November 10--14, 2026, Rio de Janeiro, Brazil}
\acmISBN{979-8-4007-2213-4/2026/11}
\acmSubmissionID{4485}
\begin{document}

\title[VoxZip: Semantic-Anchored Audio Compression]{VoxZip: Semantic-Anchored Temporal KV Cache Compression for Long-Context Audio Inference}

%% ===== Authors =====
%% IMPORTANT: names, affiliations, emails, ORCIDs MUST match the ACM rightsreview
%% form EXACTLY. Full institution name + city + country required (no abbreviations).
%% Per ACM: do NOT group/batch multiple authors into one long string.
%% Add one \author...\email...\affiliation block per author.

%% Authors 1-4 (Zhejiang University)
\author{Wenxu Jia}
\authornote{The first four authors contributed equally to this work.}
\email{jiawenxu@zju.edu.cn}
\orcid{0009-0009-3787-4143}
\affiliation{%
  \textsuperscript{1}\institution{Zhejiang University}
  \city{Hangzhou}
  \country{China}
}
\affiliation{%
  \textsuperscript{2}\institution{Meituan}
  \city{Shanghai}
  \country{China}
}

\author{Dongjie Fu}
\authornotemark[1]
\email{fudongjie@zju.edu.cn}
\orcid{0009-0000-7682-7678}
\affiliation{%
  \textsuperscript{1}\institution{Zhejiang University}
  \city{Hangzhou}
  \country{China}
}

\author{Xize Cheng}
\authornotemark[1]
\email{chengxize@zju.edu.cn}
\orcid{0000-0001-9708-3225}
\affiliation{%
  \textsuperscript{1}\institution{Zhejiang University}
  \city{Hangzhou}
  \country{China}
}

\author{Fangming Feng}
\authornotemark[1]
\email{fangmingfeng@zju.edu.cn}
\orcid{0009-0001-0814-1698}
\affiliation{%
  \textsuperscript{1}\institution{Zhejiang University}
  \city{Hangzhou}
  \country{China}
}

\author{Linjun Li}
\email{lilinjun05@meituan.com}
\orcid{0000-0003-0395-0231}
\affiliation{%
  \textsuperscript{2}\institution{Meituan}
  \city{Shanghai}
  \country{China}
}

\author{Wenshi Chen}
\email{wenshi.chen@meituan.com}
\orcid{0009-0001-4669-7873}
\affiliation{%
  \textsuperscript{2}\institution{Meituan}
  \city{Shanghai}
  \country{China}
}

\author{Yingming Li}
\email{yingming@zju.edu.cn}
\orcid{0000-0001-8011-8970}
\affiliation{%
  \textsuperscript{1}\institution{Zhejiang University}
  \city{Hangzhou}
  \country{China}
}

\author{Zhou Zhao}
\email{zhaozhou@zju.edu.cn}
\orcid{0000-0001-6121-0384}
\affiliation{%
  \textsuperscript{1}\institution{Zhejiang University}
  \city{Hangzhou}
  \country{China}
}

\author{Tao Jin}
\authornote{Corresponding author: Tao Jin (jint\_zju@zju.edu.cn).}
\email{jint\_zju@zju.edu.cn}
\orcid{0000-0003-3564-1628}
\affiliation{%
  \textsuperscript{1}\institution{Zhejiang University}
  \city{Hangzhou}
  \country{China}
}

%% Even-page header (ACM requirement: full first author name, no italics, no superscripts)
\renewcommand{\shortauthors}{Wenxu Jia et al.}


\begin{abstract}

Recent advancements in Speech Large Language Models have demonstrated remarkable capabilities in understanding complex audio tasks. Despite this progress, their long-context inference remains severely bottlenecked by prohibitive KV cache memory demands. Existing text-centric compression methods struggle here, often disrupting speech continuity or discarding crucial semantic cues. To address this, we propose VoxZip, a train-free, two-stage semantic-anchored KV cache compression framework. The first stage uses automatic speech recognition (ASR) transcriptions as explicit semantic anchors to temporally align, compress, and fuse audio tokens, significantly reducing the initial KV cache while elevating token information density. To further improve the compression ratio, the second stage employs a dynamic filtering strategy based on temporally decayed accumulated attention to evict non-essential tokens while mitigating early-token bias. Comprehensive evaluations on Qwen3-Omni across six diverse audio benchmarks demonstrate the superiority of our approach. VoxZip excels in long-audio reasoning and consistently maintains high-fidelity perception on short-form tasks. Notably, it sustains over 90\% of the uncompressed baseline performance even under an aggressive 20x KV cache compression in long-context scenarios. Furthermore, at a 4x compression ratio, VoxZip yields a 1.9x increase in inference throughput alongside a 3.3x reduction in peak memory overhead. Code and models will be available at \url{https://github.com/MM-Speech/VoxZip}.

\end{abstract}





\maketitle

\section{Introduction}

Speech Large Language Models (SLLMs) have advanced rapidly \cite{survey_sllms_1, survey_sllms_2}. The shift from cascade pipelines to unified end-to-end architectures strengthens the perception of multi-dimensional audio information, including linguistic content, acoustics, and paralinguistic cues \cite{AudioFlamingo-3,Qwen3-Omni,GLM-4-Voice}, enabling applications such as intelligent customer service \cite{fu-etal-2025-pachat} and long-form meeting abstraction. However, long-context SLLM inference remains severely bottlenecked by computational and memory costs \cite{KV_cache_expense}. Beyond model size and attention complexity, the KV cache accumulated during prefilling and generation incurs prohibitive memory overhead \cite{prompt_compression_survey_1,kv_cache_survey_1,kv_cache_survey_2}, making aggressive KV cache compression without compromising holistic understanding a critical open challenge.

\begin{figure}[htbp]
    \centering
    \includegraphics[width=\columnwidth]{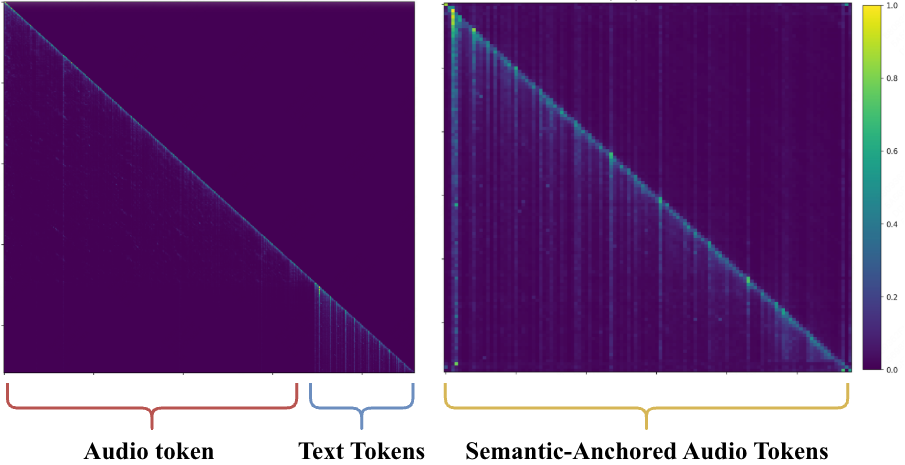}
    \caption{Attention score distributions in Qwen3-Omni-Instruct-30B. Left: the uncompressed baseline shows highly sparse attention over raw audio and text tokens. Right: semantic-anchored compression densifies and focuses the attention.}
    \label{fig:audio_attention_observation}  
\end{figure}

KV cache compression has shown strong potential in text and vision, with mainstream eviction strategies relying on accumulated attention scores or fixed windows for token selection \cite{Snapkv, SLLM, FastV, DyCoke, LOOK_M, d2o}. Yet directly applying them to Speech LLMs is difficult due to the temporal redundancy and low information density of acoustic signals. As shown in Figure \ref{fig:audio_attention_observation}, attention distributions exhibit pronounced modal heterogeneity, with textual regions yielding highly salient attention matrices, whereas raw audio tokens suffer severe attention dilution and distribution diffusion. This lack of distinctiveness makes attention-score-only KV compression unreliable for pinpointing critical semantic anchors in long-form audio, frequently causing semantic drift or hallucinations.

To address these challenges, we propose VoxZip, a train-free, two-stage compression framework that reshapes information density via semantic anchoring. In the first stage, VoxZip treats ASR-transcribed text as explicit semantic anchors and leverages their timestamps to temporally align and extract the corresponding acoustic intervals; fusing these raw audio representations with anchor embeddings condenses redundant speech into high-density semantic segments. In the second stage, a time-decayed accumulated attention score metric dynamically prunes cached tokens, mitigating the accumulation bias of early tokens to selectively safeguard essential semantic and acoustic cues. Consequently, VoxZip sustains robust linguistic comprehension and paralinguistic perception under a minimal memory footprint.

We evaluate VoxZip extensively across long-audio reasoning benchmarks (Vox-Infinity \cite{Vox-Infinity}, AudioMarathon \cite{AudioMarathon}, SPIRAL \cite{SpeechPrune}) and general audio-language benchmarks (MMSU \cite{MMSU}, MMAU \cite{MMAU}, MMAR \cite{MMAR}). Results show that VoxZip achieves a superior trade-off between perception fidelity and inference efficiency, maintaining robust performance under extreme context lengths while significantly alleviating the memory bottlenecks of SLLMs. Extended evaluations further confirm robustness against acoustic disturbances, ensuring stable inference even when environmental noise compromises the ASR-guided anchors. The main contributions are summarized below.

\begin{itemize}
    \item A semantic-anchored audio compression mechanism that compresses semantic audio segments via textual anchors, shrinking the initial KV cache to accelerate prefill while boosting semantic reasoning in long-context scenarios.

    \item A KV cache eviction strategy based on temporally decayed accumulated attention scores, which mitigates the early-token accumulation bias prevalent in long contexts and ensures high performance stability even under extreme compression constraints.

    \item Extensive evaluations across six benchmarks show VoxZip consistently outperforms baselines, retaining over 90\% performance at a 20$\times$ compression ratio, with a 1.93$\times$ throughput boost and a 3.34$\times$ reduction in peak memory usage.
\end{itemize}

\section{Related Works}

\subsection{Speech Large Language Models}

% 近年来，音频语言大模型（SLLMs）经历了从传统的“语音识别+文本LLM”模块化流水线向端到端统一架构的重大转变。以 Whisper 为代表的模型通过大规模弱监督预训练，在语义理解上奠定了坚实基础；随后，Qwen-Omni、SALMONN 以及 Gemini 1.5 Pro 等模型进一步增强了音频推理能力，使模型不仅能理解语音内容，还能感知情感、说话人身份及环境背景等副语言信息。然而，随着应用场景向长会议摘要和多轮长对话延伸，自注意力机制的平方级复杂度和 KV cache 的线性增长成为了核心瓶颈，极大地限制了长音频任务在资源受限硬件上的实时部署。

Recent research in SLLMs has driven a paradigm shift from traditional cascaded pipelines to unified end-to-end architectures \cite{survey_sllms_1, survey_sllms_2}. Foundational models such as Whisper have established robust semantic representations of speech through large-scale weakly supervised pre-training \cite{Whisper} . Subsequently, advanced frameworks such as Qwen-Omni \cite{Qwen2.5-Omni, Qwen3-Omni}, Audio Flamingo \cite{AudioFlamingo-2, AudioFlamingo-3}, and Kimi-Audio \cite{Kimi-Audio, fu2026characterspeechleveragingroleplaying, cao2026xopdcrossmodalonpolicydistillation} have further expanded the boundaries of speech reasoning. Beyond precise linguistic transcription, these models demonstrate an enhanced capability to perceive nuanced paralinguistic cues, such as emotion, speaker traits, and environmental acoustics \cite{sllm_survey_3}. However, because end-to-end models must directly process audio features that significantly longer token representations, critical system bottlenecks emerge as application scenarios scale to long-context tasks like extensive meeting summarization and ultra-multi-turn dialogues \cite{Vox-Infinity,AudioMarathon,BLAB}. The quadratic computational complexity of the self-attention mechanism, coupled with the linear memory accumulation of the Key-Value (KV) cache, imposes severe memory overheads \cite{KV_cache_expense}. Consequently, these architectural constraints significantly hinder the efficient inference and real-time deployment of long-form audio tasks on resource-constrained hardware.

\subsection{KV Cache Compression in LLMs}

%针对大规模语言模型在推理过程中的显存压力，研究界首先在文本模态提出了多种 KV Cache 压缩策略。早期的研究如 StreamingLLM 和 SnapKV 主要集中于 Token 级的 Attention Sink 机制与重要性采样；随后，PyramidKV 等工作进一步挖掘了不同网络层对上下文关注的差异性，实现了层级化压缩；而 ChunkKV 则探索了基于语义块的粗粒度压缩范式。随着多模态模型的发展，视觉领域的压缩方法也逐渐兴起。例如，FastV 和 DyCoke 针对图像与视频的时空冗余性设计了动态剪枝策略，为视觉领域的高效长序列推理提供了成熟的显存优化方案。
%然而，上述针对文本或视觉模态的压缩方法难以直接迁移至音频模态。相较于文本，音频信号具有较高的时序冗余与相对较低的单位信息密度，且缺乏明确的离散语义边界。针对音频大模型的 KV Cache 压缩，近期的工作如SpeechPrune 尝试对语音 Token 进行剪枝，但未能引入自回归解码过程中的动态压缩机制。另一项工作 \textbf{Segment-KV} 尝试通过对语音进行分段处理来压缩缓存，但其评估基准的平均时长较短，并且这种硬性划分语义段的做法容易切断连续的声学特征，导致跨片段的上下文与副语言信息流失。本文提出的 \textbf{VoxZip} 填补了这一空白。通过 ASR 引导的语义锚定音频语义段，并在解码阶段引入带有时序衰减机制的动态过滤策略以剔除低信息量token，我们在多个长音频基准测试上不仅实现了显著的压缩比，还最大程度地保留了模型原有的理解与感知性能。

To mitigate the memory overhead of large language models (LLMs) during inference, the research community initially proposed various KV cache compression strategies in the text modality \cite{kv_cache_survey_1, kv_cache_survey_2, kv_cache_review_1}. Early works such as StreamingLLM \cite{SLLM} and SnapKV \cite{Snapkv} primarily focused on token-level attention sink mechanisms and importance sampling. Subsequently, methods like PyramidKV exploited the heterogeneous attention patterns across network layers to achieve layer-wise compression \cite{pyramidKV}, while ChunkKV \cite{ChunkKV} explored a coarse-grained compression paradigm based on semantic chunks . With the rapid evolution of multimodal models, compression techniques in the vision domain have also emerged. For instance, FastV \cite{FastV} and DyCoke \cite{DyCoke} designed dynamic pruning strategies tailored to the spatial-temporal redundancy of images and videos, establishing robust memory optimization paradigms for efficient long-context visual reasoning.

However, directly transferring compression techniques from text or vision modalities to the audio domain is challenging due to the inherent temporal redundancy and low information density of acoustic signals. Currently, KV cache compression for Speech LLMs remains largely underexplored. While a recent study \cite{SpeechPrune} investigates token pruning for speech representations, it lacks a dynamic eviction mechanism during the autoregressive decoding stage, and its closed-source nature currently precludes direct empirical comparison. Our proposed VoxZip fills this critical gap. By leveraging ASR-guided semantic anchoring and introducing a dynamic filtering strategy with a temporal-decay mechanism to evict low information tokens during decoding, we achieve substantial compression ratios across multiple long-audio benchmarks while effectively preserving the model's inherent comprehension and perception capabilities.

\section{Methodology}

\subsection{Preliminaries: KV Cache in Speech LLMs}

In standard SLLMs, the generative inference process typically consists of two main stages, namely the prompt prefilling stage and the auto-regressive decoding stage. To avoid the redundant computation of historical states, SLLMs store past Key and Value representations in GPU memory, forming the KV cache.

\textbf{Prefilling stage}. Given raw audio and textual inputs, the model first extracts the corresponding acoustic and textual embeddings. These embeddings are then processed by the Transformer layers of the SLLM to compute the initial key and value matrices, denoted as $K_0$ and $V_0$. These matrices are stored to initialize the KV cache for the entire prompt sequence.

\textbf{Decoding stage}. The model generates new tokens auto-regressively. At time step $t$, for the newly generated token $x_t$, the model only computes its corresponding query $q_t$, key $k_t$, and value $v_t \in \mathbb{R}^{1 \times d}$, where $d$ is the hidden dimension. The model then retrieves the historical KV cache $K_{t-1}$ and $V_{t-1}$, dynamically updating it by concatenating the new representations:

\begin{equation}
K_t = [K_{t-1}, k_t], \quad V_t = [V_{t-1}, v_t]
\end{equation}

here $[\cdot, \cdot]$ denotes the concatenation operation along the sequence dimension. The attention output $o_t$ for the current step is subsequently computed using the updated cache:

\begin{equation}
o_t = \text{Softmax}\left(\frac{q_t K_t^\top}{\sqrt{d}}\right) V_t
\end{equation}

Consequently, the size of the KV cache grows linearly with the length of the input and generated sequences. This $\mathcal{O}(L)$ linear expansion significantly increases inference latency and memory footprint, particularly in long-context scenarios. Furthermore, unlike highly discrete text sequences, raw audio inputs exhibit extreme temporal redundancy, often producing massive audio tokens even for short utterances. Therefore, developing an efficient, modality-aware KV cache compression strategy is imperative to mitigate the memory bottleneck in SLLMs.

\begin{figure*}[htbp]
    \centering
    \includegraphics[width=\textwidth]{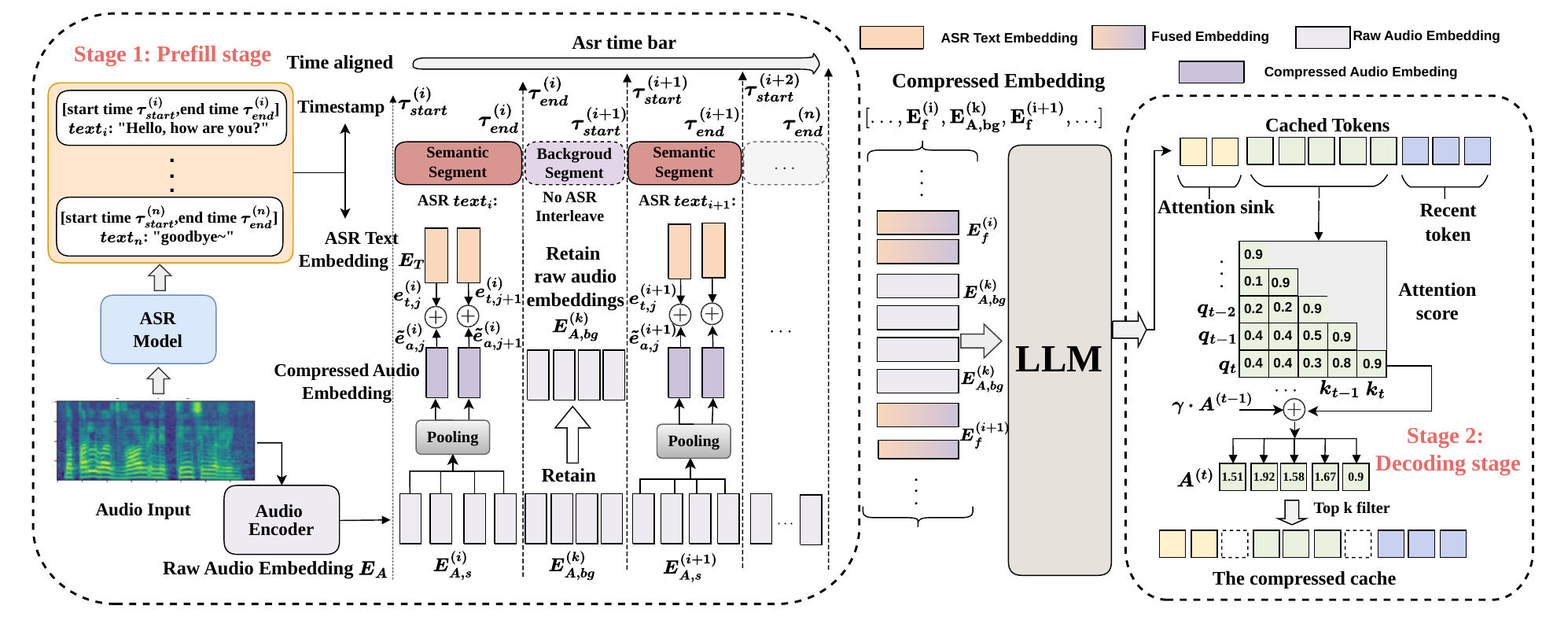}
    \caption{The VoxZip framework. Stage 1 (left) performs semantic-anchored audio compression during the prefill stage. Stage 2 (right) dynamically filters the KV cache using time-decayed attention scores during decoding.}
    \label{fig:method_overview}  
\end{figure*}

\subsection{The Proposed VoxZip}

In this section, we present VoxZip. As illustrated in Figure~\ref{fig:method_overview}, VoxZip employs a two-stage compression strategy. The first stage compresses audio tokens under the guidance of semantic anchors. The second stage further dynamically filters out low-information tokens based on temporally decayed accumulated attention scores. This approach aims to preserve model performance while reducing the token count as extensively as possible.

\subsubsection{Semantic-Anchored Audio Compression}
As shown in Figure \ref{fig:audio_attention_observation}, previous observations indicate that the attention of audio input in SLLMs suffers from attention dilution over long sequences. Unlike discrete text sequences, continuous audio streams exhibit lower information density and typically lack highly distinguishable semantic boundaries. This characteristic severely limits the performance of SLLMs in long-audio reasoning and retrieval tasks. To address this issue and enhance the model's comprehension of long contexts, we propose a semantic-anchored audio compression method. This approach utilizes transcribed text as explicit guides to compress the corresponding acoustic intervals, thereby significantly elevating the information density of these semantic regions. Specifically, the first-stage compression comprises three steps.

\textbf{Temporal-Aligned Embedding Partitioning.} Given a raw audio waveform input $A_{raw}$ with a total duration of $\tau_{total}$, we utilize an auxiliary ASR model to generate initial transcriptions. To ensure robustness against environmental noise, we only retain transcribed segments with a confidence score above a threshold of 0.3. The remaining high-confidence transcriptions constitute the valid semantic set $\mathcal{S} = \{S_1, \dots, S_n\}$, where each segment $S_i = (\tau_{start}^{(i)}, \tau_{end}^{(i)}, \text{text}^{(i)})$ contains the start time, end time, and the corresponding text. This preemptive filtering prevents erroneous semantic anchors from corrupting the continuous audio latent space. Concurrently, the SLLM encodes the raw audio $A_{raw}$ into a continuous acoustic feature sequence $E_A \in \mathbb{R}^{L_a \times d}$, while embedding each valid transcribed text $\text{text}^{(i)}$ into $E_T^{(i)} \in \mathbb{R}^{L_t^{(i)} \times d}$. Here, $L_a$ and $L_t^{(i)}$ denote their respective sequence lengths, and $d$ represents the shared hidden dimension.

Crucially, instead of physically truncating the raw audio waveform, we slice directly within the extracted continuous acoustic latent space $E_A$. Given the strong continuity of speech, this feature-level segmentation preserves the complete global acoustic context. Based on the obtained timestamps, we map the absolute time to discrete audio token indices:

\begin{equation}
    idx_{start}^{(i)} = \lfloor L_a \cdot \left(\frac{\tau_{start}^{(i)}}{\tau_{total}}\right) \rfloor, \quad idx_{end}^{(i)} = \lfloor L_a \cdot \left(\frac{\tau_{end}^{(i)}}{\tau_{total}}\right) \rfloor
\end{equation}

Using these indices, we slice $E_A$ to extract the \textit{semantic audio intervals}, denoted as $E_{A,s}^{(i)} = E_A[idx_{start}^{(i)} : idx_{end}^{(i)}]$. The remaining parts of the sequence, which lack explicit semantic boundaries and are not covered by any transcribed intervals, are identified as \textit{background acoustic intervals} and denoted as $\{E_{A,bg}^{(1)}, \dots, E_{A,bg}^{(m)}\}$.

\textbf{Semantic-Anchored Audio Compression.} For the $i$-th semantic audio interval, let $E_{A,s}^{(i)} = [e_{a,1}^{(i)}, \dots, e_{a,L_a^{(i)}}^{(i)}] \in \mathbb{R}^{L_a^{(i)} \times d}$ denote its acoustic feature sequence, and let $E_T^{(i)} = [e_{t,1}^{(i)}, \dots, e_{t,L_t^{(i)}}^{(i)}] \in \mathbb{R}^{L_t^{(i)} \times d}$ denote its corresponding text feature sequence. Since the audio sequence length $L_a^{(i)}$ is typically much larger than the text sequence length $L_t^{(i)}$, we uniformly partition the $L_a^{(i)}$ acoustic features into $L_t^{(i)}$ consecutive groups along the temporal dimension. For the $j$-th group $\mathcal{G}_{i,j}$, we perform an element-wise average pooling to obtain the compressed acoustic representation $\tilde{e}_{a,j}^{(i)}$:

\begin{equation}
    \tilde{e}_{a,j}^{(i)} = \frac{1}{|\mathcal{G}_{i,j}|} \sum_{k \in \mathcal{G}_{i,j}} e_{a,k}^{(i)}
\end{equation}

where $|\mathcal{G}_{i,j}|$ is the number of audio tokens in the group. Subsequently, we perform an element-wise addition between this compressed acoustic representation $\tilde{e}_{a,j}^{(i)}$ and its aligned text embedding $e_{t,j}^{(i)}$ to construct the semantic anchor $e_{f,j}^{(i)} = \tilde{e}_{a,j}^{(i)} + e_{t,j}^{(i)}$. This operation compresses the audio semantic interval into a high-density sequence $E_f^{(i)} = [e_{f,1}^{(i)}, \dots, e_{f,L_t^{(i)}}^{(i)}] \in \mathbb{R}^{L_t^{(i)} \times d}$, naturally preserving crucial paralinguistic cues while tightly aligning with core textual semantics.

\textbf{Background Preservation and Sequence Concatenation.} For the background acoustic intervals $\{E_{A,bg}^{(1)}, \dots, E_{A,bg}^{(m)}\}$, we preserve their original acoustic features to maintain the model's perception of global environmental sounds. Finally, we concatenate all the fused semantic embedding intervals $E_f^{(i)}$ and the preserved background embedding intervals $E_{A,bg}^{(k)}$ in their original chronological order to form the final compressed input sequence $E_{comp}$:

\begin{equation}
    E_{comp} = \text{Concat}(\dots, E_f^{(i)}, E_{A,bg}^{(k)}, E_f^{(i+1)}, \dots)
\end{equation}

Through this strategy, the length of the transcribed audio intervals is compressed to approximately $0.25\times$ of their original size. This significant reduction accelerates the prefill stage and substantially reduces the memory footprint of the initial KV cache. Furthermore, the introduction of semantic anchors effectively enhances the model's semantic reasoning and retrieval capabilities in long-audio contexts.

\subsubsection{Temporally Decayed KV Cache Eviction}

Following the first stage, the KV cache comprises both high-density semantic tokens and raw acoustic tokens. To further prune redundant and low information elements, we propose a dynamic KV cache eviction mechanism during the generation stage.
Mainstream score-based eviction methods typically evaluate token importance via simple accumulation of historical attention scores \cite{Snapkv,pyramidKV,H2O}. However, directly applying this to long-form audio induces a severe accumulation bias. Since early tokens exist for a longer duration and undergo more accumulation steps, they inherently amass higher scores, causing the eviction mechanism to persistently select and retain them. To mitigate this bias and preserve paralinguistic sensitivity, we introduce a temporal decay mechanism.

Specifically, operating independently per layer, the time-decayed accumulated score vector $A^{(t)} \in \mathbb{R}^{1 \times t}$ is dynamically updated using the $t$-th query $q_t \in \mathbb{R}^{1 \times d}$ and the current key cache $K_t \in \mathbb{R}^{t \times d}$:

\begin{equation}
    A^{(t)}=\gamma \cdot [A^{(t-1)}, 0] + \text{Softmax}\left(\frac{q_t K_t^\top}{\sqrt{d}}\right)
\end{equation}

where $d$ represents the head dimension, $\gamma \in (0, 1)$ is a predefined decay coefficient, and $[\cdot, \cdot]$ denotes zero-padding concatenation to align the dimension for the newly generated token. This formulation evaluates the global historical contribution of tokens while explicitly penalizing stale acoustic states.

During eviction, the cache is partitioned into an attention sink window, a recent window, and important historical tokens. To maintain generation stability, we unconditionally preserve the first $T$ attention sink tokens and the latest $M$ recent tokens. For the intermediate historical tokens, we extract the top-$N$ elements based on their decayed accumulated score $A^{(t)}$. In our implementation, we empirically set $T = 4$ and dynamically allocate 20\% of the total KV cache budget to the recent window $M$, assigning the remaining capacity to $N$. Finally, the updated KV cache is constructed via concatenation:

\begin{equation}
K_{t+1}=\text{Concat}(K_t[:T, :], K_t[\mathcal{I}_{hist}, :], K_t[-M:, :])    
\end{equation}

\begin{equation}
    V_{t+1}=\text{Concat}(V_t[:T, :], V_t[\mathcal{I}_{hist}, :], V_t[-M:, :])
\end{equation}

\begin{equation}
    \mathcal{I}_{hist}=\text{Top}_N \left(A^{(t)}[T : -M]\right)
\end{equation}

\begin{table*}[htpb]
\centering
\caption{Performance on long-context audio benchmarks under varying KV cache budgets (\%). Backbone: Qwen3-Omni-Instruct-30B. Subsets: Beyond-Semantics Dialogue (BS), Conversational Dialogues (Conv), Ultra Multi-Turn Dialogue (UMT), Personal Monologue (PM), AudioMarathon (SCE: Speech Content Extraction; AC: Audio Classification; SR: Speaker Recognition), and SPIRAL Hard (H). Aver.: average accuracy. PRR: performance retention ratio vs.\ the Full KV baseline (\%). Bold: best within each budget.}
\label{tab:kv_compression_instruct}
\begin{tabular}{l c | cccc | ccc | c | c c}
\toprule
\multirow{2}{*}{\textbf{Method}} & \multirow{2}{*}{\textbf{Ratio}} & \multicolumn{4}{c|}{\textbf{Vox-Infinity}} & \multicolumn{3}{c|}{\textbf{AudioMarathon}} & \textbf{SPIRAL} & \multirow{2}{*}{\textbf{Aver.}} & \multirow{2}{*}{\textbf{PRR}} \\
\cmidrule(lr){3-6} \cmidrule(lr){7-9} \cmidrule(lr){10-10}
& & \textbf{BS} & \textbf{Conv} & \textbf{UMT} & \textbf{PM} & \textbf{SCE} & \textbf{AC} & \textbf{SR} & \textbf{(H)} & & \\ 
\midrule
Full KV & 100\% & 31.60 & 95.60 & 69.20 & 71.40 & 49.03 & 54.82 & 66.90 & 98.00 & 67.07 & 100.00 \\
\midrule
StreamingLLM & 25\% & \textbf{28.40} & 70.00 & 24.00 & 12.20 & 40.15 & 50.56 & 52.49 & 61.10 & 42.36 & 63.16 \\
SnapKV       & 25\% & 24.30 & 78.40 & 37.00 & 10.70 & 42.54 & 47.00 & 51.28 & 51.63 & 42.74 & 63.72 \\
PyramidKV    & 25\% & 24.80 & 85.50 & 34.40 & 11.30 & 44.15 & 47.90 & 51.73 & 58.12 & 44.43 & 66.70 \\
ChunkKV      & 25\% & 23.10 & 77.20 & 32.90 & 10.80 & 42.67 & 49.72 & 50.51 & 48.63 & 41.94 & 62.54 \\
\textbf{VoxZip (Ours)}& 25\% & 26.60 & \textbf{96.20} & \textbf{82.40} & \textbf{86.30} & \textbf{55.94} & \textbf{62.89} & \textbf{54.63} & \textbf{93.02} & \textbf{69.75} & \textbf{104.00} \\
\midrule
StreamingLLM & 15\% & \textbf{28.00} & 67.60 & 18.60 & 8.60  & 39.83 & 41.98 & 48.32 & 53.38 & 38.28 & 57.08 \\
SnapKV       & 15\% & 22.00 & 71.20 & 19.80 & 8.80  & 41.36 & 40.67 & 52.18 & 46.38 & 37.00 & 55.16 \\
PyramidKV    & 15\% & 25.20 & 80.60 & 15.00 & 10.30 & 37.18 & 37.27 & 45.48 & 47.38 & 37.27 & 55.57 \\
ChunkKV      & 15\% & 22.80 & 72.00 & 20.60 & 7.20  & 42.87 & 45.73 & 49.75 & 47.63 & 38.57 & 57.51 \\
\textbf{VoxZip (Ours)}& 15\% & 24.10 & \textbf{98.00} & \textbf{67.20} & \textbf{81.60} & \textbf{56.28} & \textbf{62.15} & \textbf{54.07} & \textbf{93.72} & \textbf{67.14} & \textbf{100.10} \\
\midrule
\textbf{VoxZip (Ours)}& 10\% & 23.80 & 98.10 & 55.00 & 76.50 & 55.07 & 60.63 & 53.00 & 94.27 & 64.55 & 96.24 \\ 
\textbf{VoxZip (Ours)}& 5\%  & 22.00 & 97.60 & 37.00 & 70.10 & 55.33 & 62.13 & 53.31 & 93.02 & 61.31 & 91.41 \\ 
\bottomrule
\end{tabular}
\end{table*}

where $\mathcal{I}_{hist}$ represents the corresponding indices of the selected tokens in the original sequence. This dynamic mechanism strictly retains only the most essential semantic and acoustic tokens to bound the inference context. Consequently, it achieves a high compression ratio without sacrificing model performance.

\section{Experiments}
\subsection{Experimental Setup and Implementation Details}

\subsubsection{Benchmark}

To evaluate VoxZip, we select two categories of benchmarks. The first category, Long-Context Reasoning and Retrieval, is headlined by Vox-Infinity \cite{Vox-Infinity}, the most demanding benchmark requiring complex multi-turn semantic retrieval and reasoning across ultra-long audio, with its Personal-Monologue subset averaging over 20 minutes. This category also includes AudioMarathon \cite{AudioMarathon} and the SPIRAL hard subset \cite{SpeechPrune}, which focus on relatively straightforward single-turn long-audio understanding. The second category, General Audio Understanding, comprises MMSU 
\cite{MMSU}, MMAU \cite{MMAU}, and MMAR \cite{MMAR}, primarily utilized to assess the model's perception of fine-grained acoustic features and paralinguistic cues in shorter contexts.

\begin{table*}[t]
\centering
\caption{Performance on general audio benchmarks (Qwen3-Omni-Instruct-30B). \textbf{Ratio}: KV cache budget (\%). \textbf{MMAR}: Semantic, Cultural, Perception, Signal; \textbf{MMSU}: Semantics, Phonology, Style, Traits; \textbf{MMAU}: Sound, Speech, Music. \textbf{Aver.}: average accuracy. \textbf{PRR}: performance retention ratio vs.\ Full KV (\%).}
\label{tab:short_audio_bench_kv_instruct_only}
\begin{tabular}{l c | cccc | cccc | ccc | c | c}
\toprule
\multirow{2}{*}{\textbf{Method}} & \multirow{2}{*}{\textbf{Ratio}} & \multicolumn{4}{c|}{\textbf{MMAR}} & \multicolumn{4}{c|}{\textbf{MMSU}} & \multicolumn{3}{c|}{\textbf{MMAU}} & \multirow{2}{*}{\textbf{Aver.}} & \multirow{2}{*}{\textbf{PRR}} \\
\cmidrule(lr){3-6} \cmidrule(lr){7-10} \cmidrule(lr){11-13}
& & \textbf{Sem.} & \textbf{Cul.} & \textbf{Per.} & \textbf{Sig.} & \textbf{Sem.} & \textbf{Pho.} & \textbf{Sty.} & \textbf{Tra.} & \textbf{Snd.} & \textbf{Spe.} & \textbf{Mus.} & & \\ 
\midrule
Full KV   & 100\% & 70.56 & 58.39 & 55.45 & 46.51 & 81.64 & \textbf{69.97}& 54.03 & 35.35 & 80.48 & 78.08 & 74.25 & 65.18 & 100.0\\
\midrule
StreamingLLM     & 25\%  & 47.69 & 40.15 & 38.12 & 34.88 & 57.86 & 54.44 & 51.48 & 35.48 & 40.54 & 36.23 & 29.43 & 42.39 & 65.04\\
SnapKV    & 25\%  & 39.17 & 46.72 & 37.13 & 48.84 & 38.45 & 53.61 & 40.43 & 35.48 & 25.23 & 23.63 & 22.52 & 37.35 & 57.30\\
PyramidKV & 25\%  & 55.47 & 60.58 & 48.51 & 55.81 & 51.79 & \textbf{72.92} & 54.49 & 42.61 & 71.17 & 65.87 & 58.86 & 58.01 & 89.01\\
ChunkKV   & 25\%  & 44.53 & 50.36 & 45.30 & \textbf{69.77} & 38.65 & 60.90 & \textbf{61.11} & \textbf{44.27} & 74.47 & 64.07 & 50.45 & 54.91 & 84.24\\
\textbf{VoxZip (Ours)}& 25\% & \textbf{72.51} & \textbf{72.99}& \textbf{67.24}& 61.54& \textbf{72.35} & 63.72& 56.80& 37.53& \textbf{80.78}& 75.48& \textbf{75.45}& \textbf{66.93} & \textbf{102.64}\\
\midrule
\textbf{VoxZip (Ours)}& 15\% & 72.75 & 70.80 & 66.09 & 58.14 & 72.49& 65.06& 56.74& 37.53& 80.48 & 75.45 & 75.77 & 66.48& 102.01\\
\textbf{VoxZip (Ours)}& 10\% & 72.51 & 69.34 & 65.84 & 58.14 & 70.06& 62.04& 56.74& 37.66& 79.88 & 74.31 & 75.77 & 65.66& 100.86\\
\bottomrule
\end{tabular}
\end{table*}

\subsubsection{Baselines}
We evaluated our approach against five baselines. Full KV serves as the uncompressed baseline, which retains the complete KV cache. For compressed baselines, we include SLLM \cite{SLLM}, which maintains performance by preserving attention sink tokens and the most recent tokens. We further evaluate SnapKV \cite{Snapkv}, a widely adopted method that selects the top-$k$ tokens based on accumulated attention scores. To capture more sophisticated compression paradigms, we introduced PyramidKV \cite{pyramidKV}, which allocates KV cache budgets hierarchically across different layers, and ChunkKV \cite{ChunkKV}, which retains the cache based on semantic segments.

\subsubsection{Implementation Details}
%所有实验均在NVIDIA A100 80GB GPU上进行
All experiments are conducted on a single computing node equipped with 8 NVIDIA A100 (80GB) GPUs. For the ASR model configuration, we utilize Whisper-Turbo as the ASR model and Qwen3-Omni-30B (both Instruct and Thinking variants) as the backbone LLM. For the hyperparameter settings, the temporal decay factor $\gamma$ used for attention score aggregation in Stage 2 is empirically set to 0.95.

\subsection{Performance on Long-Context Audio Benchmarks}

To evaluate the long-context retrieval and reasoning capabilities of various compression algorithms, we evaluate the Qwen3-Omni-Instruct-30B and Qwen3-Omni-Thinking-30B on Vox-Infinity, AudioMarathon, and the SPIRAL (Hard) subset. These benchmarks rigorously test the model's ability to extract core semantic features under substantial KV cache constraints.

As shown in Table \ref{tab:kv_compression_instruct}, at a 25\% cache budget, VoxZip consistently outperforms all baselines and notably surpasses the uncompressed Full KV baseline in average accuracy. This gain is fundamentally driven by introducing ASR transcriptions as explicit semantic anchors. Fusing these condensed textual features with aligned audio tokens reconstructs the sequence's information density. These injected textual priors provide reliable guidance, significantly enhancing long-audio semantic understanding while preserving essential paralinguistic cues.

VoxZip maintains robust performance even under extreme compression. At a 10\% cache budget, both the Instruct and Thinking variants retain over 90\% of their uncompressed baseline, demonstrating the method's robust generalization across distinct tuning architectures. Even at an extreme 5\% KV cache budget, VoxZip sustains a 91.41\% performance retention, significantly outperforming the strongest baseline PyramidKV \cite{pyramidKV}, which retains only 66.70\% at a much looser 25\% budget. This resilience stems from the framework's synergistic design, guided by the first-stage textual semantic anchors, with the second stage employing a temporally decayed accumulated attention mechanism to dynamically evict low-information tokens. This formulation effectively mitigates early-token bias, enabling aggressive pruning without compromising critical semantic and paralinguistic cues.

\subsection{Performance on General Audio QA Benchmarks}
%接下来，我们选择了三个侧重于声学和副语言信息理解的音频qa数据集，与前者不同的是，这些数据集的平均时长都在30s以下。该实验主要是用来测量我们模型压缩后对声学的保留

Beyond long-range retrieval, we evaluate the acoustic perception precision of Qwen3-Omni-Instruct-30B on general audio question-answering (QA) benchmarks including MMAR \cite{MMAR}, MMSU \cite{MMSU} and MMAU \cite{MMAU}. Unlike long-audio tasks, these focus on fine-grained features within a 30-second window, such as speaker traits, emotional and prosodic nuances, and low-level signal characteristics. This critically assesses whether VoxZip preserves the rich non-linguistic cues inherently absent from discrete text transcriptions.

As illustrated in Table \ref{tab:short_audio_bench_kv_instruct_only}, under a 25\% KV cache budget ratio, VoxZip achieves an average accuracy of 66.93\%, surpassing the Full KV baseline by 1.75\%. By contrast, competing KV compression schemes suffer notable degradation, with even the strongest PyramidKV baseline reaching only 58.01\% average accuracy. Crucially, VoxZip maintains its superiority under extreme compression by sustaining average accuracies of 66.48\% and 65.66\% at stricter 15\% and 10\% budgets respectively, thereby consistently outperforming the uncompressed Full KV baseline. Furthermore, VoxZip demonstrates extraordinary proficiency in tasks reliant on complex acoustic profiling. Specifically, on the Perception and Signal subsets of MMAR, it achieves accuracy gains of over 10\% against Full KV.

We attribute these performance gains over the uncompressed baseline to our framework's synergistic design. The first-stage fusion robustly preserves essential paralinguistic cues, while the second-stage dynamic filtering evicts redundant tokens to suppress irrelevant acoustic noise. This denoised representation enables the model to focus precisely on salient acoustic events, yielding superior accuracy under a constrained token budget.

\subsection{Ablation Studies}

\subsubsection{Necessity of Semantic Anchors in Long Audio Understanding}

To validate the necessity of semantic anchors in long audio understanding, we conducted an ablation study on three semantic-focused subsets of the Vox-Infinity benchmark. This dataset was specifically selected as its ultra-long context rigorously tests the model's ability to accurately retrieve critical semantics amidst massive acoustic noise.

As shown in Table \ref{tab:ablation_semantic_single}, removing the semantic-anchored compression (w/o Semantic-Anchored Comp.) at a 15\% budget leads to severe performance degradation, with accuracy dropping to 6.4\% on the longest PM subset. In contrast, integrating the semantic anchors (w/ Semantic-Anchored Comp.) not only prevents this collapse by boosting the PM performance to an exceptional 81.6\%, but also elevates the overall average to 82.27\%, directly surpassing the uncompressed Full KV baseline of 78.7\%.

This capability to exceed the uncompressed upper bound fundamentally stems from the reconstruction of information density facilitated by semantic anchors. As illustrated by the attention maps in Figure \ref{fig:audio_attention_observation}, under unanchored pure audio conditions, the model's attention toward low-density audio tokens is highly discrete and sparse. This sparsity leads to a severe loss of focus during long-context reasoning. Conversely, introducing semantic anchors injects highly condensed semantic priors into the continuous audio stream, densifying the originally scattered attention scores. This mechanism effectively guides the model to bypass acoustic redundancy and concentrate strictly on anchor segments that encapsulate core semantics. Consequently, it fortifies the robustness of long-range semantic retrieval and provides highly reliable scoring criteria for the subsequent Stage-2 KV cache eviction.

\begin{table}[t]
\centering
\caption{Ablation of semantic-anchored audio compression on Vox-Infinity semantic subsets. \textbf{w/}: with; \textbf{w/o}: without; \textbf{Comp.}: compression. Accuracy (\%).}
\label{tab:ablation_semantic_single}
\begin{tabular}{l | ccc | c}
\toprule
\textbf{Method} & \textbf{Conv} & \textbf{UMT} & \textbf{PM} & \textbf{Avg.} \\ 
\midrule
Full KV & 95.60 & 69.20 & 71.40 & 78.70 \\
\midrule
\rowcolor{gray!5} \multicolumn{5}{l}{\textit{KV Cache Budget Ratio: 15\%}} \\
w/o Semantic-Anchored Comp. & 87.04 & 44.80 & 6.40 & 46.08 \\
w/ Semantic-Anchored Comp. & \textbf{98.00} & \textbf{67.20} & \textbf{81.60} & \textbf{82.27} \\
\midrule
\rowcolor{gray!5} \multicolumn{5}{l}{\textit{KV Cache Budget Ratio: 10\%}} \\
w/o Semantic-Anchored Comp. & 85.80 & 36.40 & 7.51 & 43.24 \\
w/ Semantic-Anchored Comp. & \textbf{98.10} & \textbf{55.00} & \textbf{76.50} & \textbf{76.53} \\
\bottomrule
\end{tabular}
\end{table}

\begin{table*}[t]
\centering
\caption{Ablation of Stage-1 fusion strategies. Accuracy (\%). Configurations: (1) T (Text-only): ASR transcriptions only; (2) A (Audio-only): raw audio only; (3) T+A (Partial): speech segments only, background discarded; (4) T+A (Full): our standard fusion. Acoustic Gain: absolute gain of T+A (Full) over T (Text-only).}
\label{tab:ablation_complete_english}
\begin{tabular}{l | ccc | ccc | ccc | cc | c | c} 
\toprule
\multirow{2}{*}{\textbf{Configuration}} & \multicolumn{3}{c|}{\textbf{MMAR (Layer-wise)}} & \multicolumn{3}{c|}{\textbf{MMSU (Acoustic/Trait)}} & \multicolumn{3}{c|}{\textbf{MMAU (Task)}} & \multicolumn{2}{c|}{\textbf{AM (Subset)}} & \textbf{Vox} & \multirow{2}{*}{\textbf{Avg.}} \\
\cmidrule(lr){2-4} \cmidrule(lr){5-7} \cmidrule(lr){8-10} \cmidrule(lr){11-12} \cmidrule(lr){13-13}
& Sign. & Perc. & Cult. & Phon. & Style & Trait & Snd. & Mus. & Spch. & SR & AC & BS & \\
\midrule
(1) T (Text-only) & 46.51 & 55.45 & 58.39 & 60.27 & 47.28 & 30.46 & 65.47 & 61.38 & 75.38 & 61.34 & 52.25 & 21.80 & 53.00 \\
(2) A (Audio-only) & 69.77 & 69.31 & 73.72 & 69.97 & 54.03 & 35.35 & 80.48 & 74.25 & 78.08 & 50.56 & 66.90 & 31.60 & 62.84 \\
\midrule
(3) T+A (Partial) & 55.81 & 59.90 & 67.88 & 60.06 & 48.70 & 43.06 & 78.39 & 73.05 & 71.17 & 53.47 & 53.67 & 23.80 & 57.41 \\
(4) \textbf{T+A (Full)} & 61.54 & 67.24 & 72.99 & 63.72 & 56.74 & 37.53 & 80.78 & 75.45 & 75.48 &  62.89 & 54.63 & 25.20 & 60.76 \\
\midrule
\rowcolor[HTML]{EFEFEF} 
\textbf{Acoustic Gain} & \textbf{+15.03} & \textbf{+11.79} & \textbf{+9.47} & \textbf{+3.50} & \textbf{+9.53} & \textbf{+7.07} & \textbf{+15.31} & \textbf{+14.10} & \textbf{+0.10} & \textbf{+1.55} & \textbf{+2.40} & \textbf{+3.40} & \textbf{+7.80} \\
\bottomrule
\end{tabular}
\end{table*}

\subsubsection{Necessity of Preserving Acoustic and Paralinguistic Cues}
To validate the design rationale of our first-stage semantic anchored compression, we conducted an ablation study across multiple benchmarks emphasizing acoustic and paralinguistic comprehension. Specifically, this study evaluates its performance advantages over the transcribed text-only baseline and the necessity of retaining un-transcribed pure audio segments during fusion. Table \ref{tab:ablation_complete_english} presents a comprehensive performance comparison of the four evaluated methods.

Fundamentally, the experimental results directly expose the inherent modality deficit of the Text-only baseline when processing complex audio tasks. While ASR models accurately extract semantic content, they inevitably strip away crucial paralinguistic cues such as emotion, speaker style, and environmental sounds. By contrast, our T+A (Full) strategy seamlessly integrates semantic and acoustic features, achieving an absolute average improvement of 7.80\% over the text-only baseline. Notably, in tasks heavily reliant on low-level acoustic perception, such as the Sign subset in MMAR and the Snd subset in MMAU, our method delivers absolute gains of 15.03\% and 15.31\% over the pure text baseline, respectively. This strongly demonstrates that utilizing ASR transcriptions as anchors to compress and fuse audio features successfully preserves rich paralinguistic information while maintaining high semantic fidelity.

Furthermore, the ablation validates the necessity of retaining the un-transcribed audio segments. Under the T+A (Partial) configuration, which aggressively discards segments with no ASR output or low transcription confidence, the average accuracy drops from 60.76\% to 57.41\%. This degradation indicates that these non-semantic audio intervals encode rich acoustic information and paralinguistic features essential for comprehensive scene and style reasoning. By continuously preserving these untranscribed or low-confidence segments, our T+A (Full) approach fully captures these underlying acoustic cues, effectively preventing the loss of acoustic information that is typically ignored by ASR transcriptions.

Finally, we reference the uncompressed Audio-only input as the theoretical performance upper bound. Although the Audio-only baseline preserves the most comprehensive information and achieves the highest average accuracy of 62.84\%, it concurrently incurs prohibitive KV cache overhead and inference latency. Encouragingly, our T+A (Full) approach maintains an exceptional performance of 60.76\% while compressing massive volumes of redundant speech frames down to the text-token length level. This strongly suggests that our proposed semantic-anchored fusion strategy achieves a highly optimal balance between efficiency and accuracy, trading an exceptionally massive token compression rate for a negligible performance degradation.

\begin{table}[t]
    \centering
    \caption{Sensitivity to the temporal decay factor ($\gamma$) on Vox-Infinity under a 5\% KV cache budget. Accuracy (\%). $\gamma = 1.0$ denotes the baseline without temporal decay. \textbf{Bold}: best.}
    \label{tab:ablation_decay}
    \begin{tabular}{lccccc}
        \toprule
        \textbf{Decay Factor ($\gamma$)} & \textbf{PM} & \textbf{UMT} & \textbf{Conv} & \textbf{BS} & \textbf{Avg.} \\
        \midrule
        $\gamma = 0.80$ & 68.40 & 35.10 & 96.90 & 20.30 & 55.17 \\
        $\gamma = 0.90$ & 73.20 & 36.40 & 97.20 & 21.40 & 57.05 \\
        \textbf{$\gamma = 0.95$} & \textbf{76.50} & 37.00 & \textbf{97.60} & \textbf{22.00} & \textbf{58.28} \\
        $\gamma = 0.99$ & 67.10 & 37.50 & 97.00 & 20.60 & 55.55 \\
        $\gamma = 1.00$ & 62.60 & \textbf{37.80} & 96.80 & 19.80 & 54.25 \\
        \bottomrule
    \end{tabular}
\end{table}

\subsubsection{Necessity of the Temporal Decay Mechanism}

To validate the temporal decay (TD) mechanism, we ablate it on Qwen3-Omni-Instruct-30B over Vox-Infinity under a highly constrained 5\% KV cache budget, comparing TD against the standard accumulation baseline ($\gamma=1.0$) and sweeping the decay factor $\gamma$. As shown in Table \ref{tab:ablation_decay}, TD yields the largest gains on the Personal Monologues (PM) subset, whose long duration demands extended temporal reasoning. The non-TD baseline's PM accuracy plummets to 62.6\%, while the optimal TD setting ($\gamma=0.95$) raises it to 76.5\%, a 13.9\% absolute improvement.

Sweeping $\gamma$ reveals an inverted-U trend: aggressive decay ($\gamma \le 0.90$) prematurely discards historical semantic anchors, while excessively slow decay ($\gamma \ge 0.99$) under-penalizes stale audio tokens, with $\gamma=0.95$ striking the optimal balance. By explicitly penalizing early tokens, TD mitigates the accumulation bias that plagues conventional score-based eviction, preserving a vital long-term perspective for token selection and reliably safeguarding reasoning stability under the extreme 5\% budget.

\begin{table}[t]
    \centering
    \caption{Efficiency at 64K context, 300 output tokens, 25\% budget, on 8$\times$A100 (80GB). \textbf{Mem} (peak memory) and \textbf{TPS} (tokens per second) include ASR overhead; Prefill is the LLM prefill latency. \textbf{Bold}: best.}
    \label{tab:efficiency}
    \begin{tabular}{lccc}
        \toprule
        \textbf{Method} & \textbf{Prefill (s)} $\downarrow$ & \textbf{Mem (GB)} $\downarrow$ & \textbf{TPS (Speedup)} $\uparrow$ \\
        \midrule
        Full KV  & 2.26 & 235.93 & 2.10 ($1.00\times$) \\
        \midrule
        SnapKV  & 1.62 & 80.04 & 3.23 ($1.54\times$) \\
        PyramidKV  & 1.66 & 89.66 & 2.87 ($1.37\times$) \\
        ChunkKV   & 1.67 & 79.98 & 3.52 ($1.68\times$) \\
        \midrule
        \textbf{VoxZip (Ours)}  & \textbf{1.30} & \textbf{70.68}& \textbf{4.06 ($\mathbf{1.93\times}$)} \\
        \bottomrule
    \end{tabular}
\end{table}

\begin{figure}[t]
    \centering
    \includegraphics[width=\linewidth]{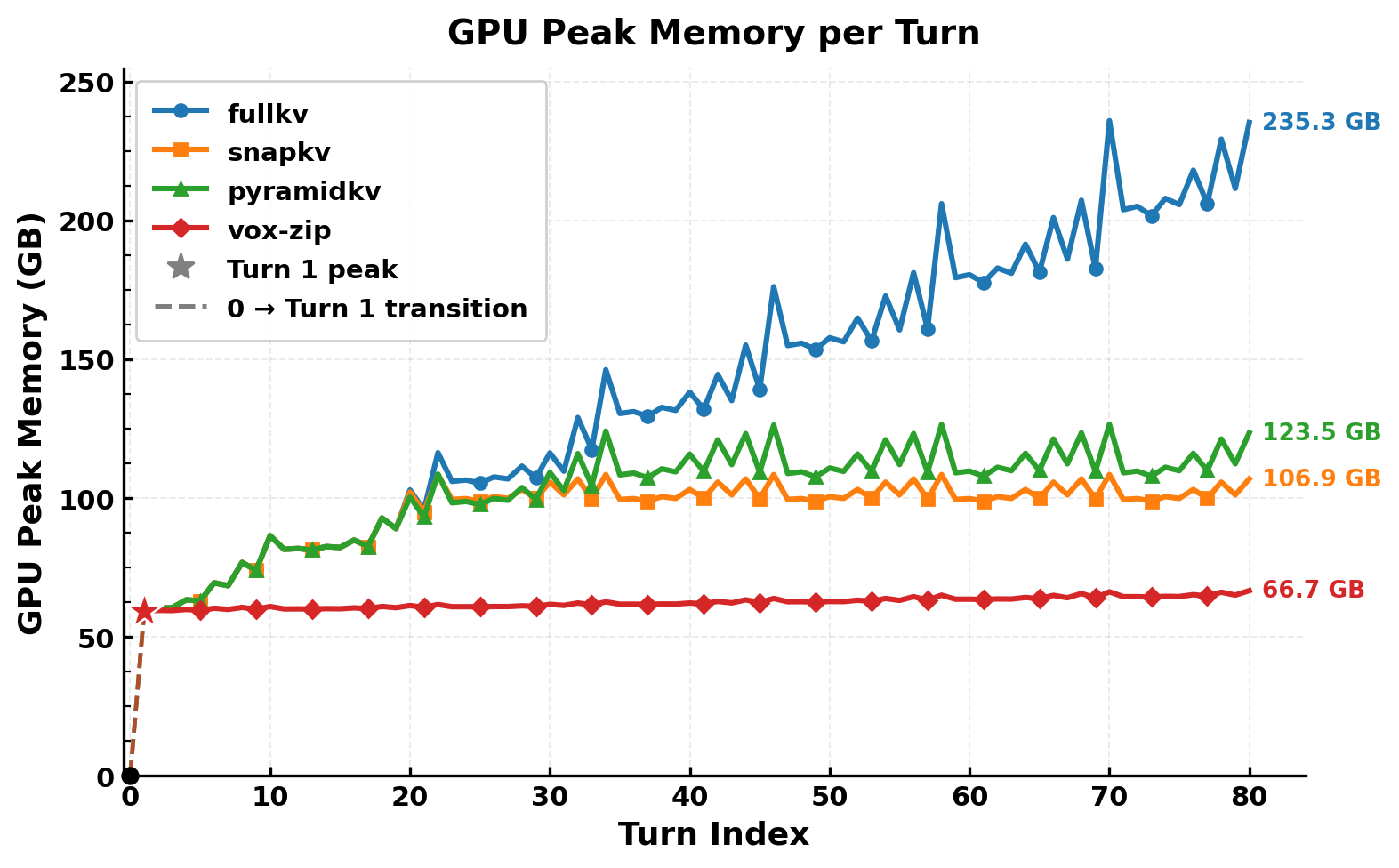}
    \caption{GPU peak memory across conversational turns.}
    \label{fig:placeholder}
\end{figure}

\subsection{Inference Efficiency Analysis}

To assess the practical computational efficiency of various compression strategies, we evaluate the Qwen3-Omni-30B-Thinking under long-context multi-turn conversational scenarios. Table \ref{tab:efficiency} details the inference metrics under a fixed 64K context window, whereas Figure \ref{fig:placeholder} tracks the dynamic peak memory accumulation across 80 continuous conversational turns. Crucially, our reported peak memory (Mem) and throughput (TPS) are based on an end-to-end measurement protocol that explicitly accounts for the auxiliary ASR module's overhead, including weight occupancy, initialization latency, and real-time transcription costs.

Experimental results demonstrate that VoxZip achieves optimal inference efficiency across the evaluated dimensions. As shown in Table \ref{tab:efficiency}, in the fixed 64K context evaluation, VoxZip exhibits superior memory efficiency, requiring a peak memory of 70.68 GB, which yields a 3.34$\times$ memory reduction compared to the Full Cache baseline. Furthermore, benefiting from the first-stage semantic anchored compression, the initial KV cache of semantic audio intervals is reduced to approximately 25\% of its original capacity. This substantial compression significantly accelerates the prefill process, bringing the latency down to 1.30 seconds (a 1.74$\times$ speedup) and thereby outperforming mainstream methods such as SnapKV and PyramidKV.

The dynamic memory profile during multi-turn conversation (Figure \ref{fig:placeholder}) further illustrates the architectural advantages of VoxZip. As conversational turns accumulate, the memory footprint of Full Cache exhibits linear growth, eventually leading to potential out-of-memory risks. Although SnapKV and PyramidKV stabilize in later turns, their lack of token compression during the prefill stage results in notable peak memory overhead due to the large volume of initial input tokens. In contrast, VoxZip reduces the sequence length directly at the input source via preemptive semantic-anchored compression. Combined with the second-stage dynamic compression mechanism that further condenses historical tokens, VoxZip maintains a gradual peak memory growth, ultimately plateauing at approximately 66 GB. This approach effectively lowers the hardware memory requirements for sustained multi-turn conversation.

\section{Limitations}
The proposed VoxZip has two main limitations. Firstly, its reliance on ASR transcriptions as semantic anchors introduces vulnerability to noise and heavy accents; while our confidence-filtering strategy mitigates this, it cannot completely eliminate the impact of low-quality transcriptions. Secondly, the element-wise addition used for audio-text fusion is inherently simplistic and may not adequately model complex cross-modal dynamics, suggesting that integrating learnable fusion mechanisms could offer substantial future improvements.

\section{Conclusion}
In this paper, we investigate the modal heterogeneity phenomenon where audio attention scores exhibit low information density relative to text. We propose VoxZip, a train-free audio KV cache compression framework that accelerates prefill and decoding. VoxZip operates in two stages: in the first stage, we use ASR transcriptions as semantic anchors to compress corresponding audio intervals and fuse them with text embeddings, increasing the information density of the audio semantic segments. In the second stage, we filter out low-information tokens by employing temporally decayed accumulated attention scores as a contribution metric, which further reduces low-density tokens and raises the overall compression ratio. Benchmark results on the Qwen3-Omni architecture across six audio tasks demonstrate that VoxZip consistently preserves holistic perception under high compression ratios. Notably, our method retains over 90\% of the original performance under a 20$\times$ KV cache compression ratio. At a 4$\times$ KV cache compression ratio, we achieve up to a 3.3$\times$ peak memory reduction and a 1.9$\times$ inference speedup.

\begin{acks}
This work was supported by the National Natural Science Foundation of China under Grant No. U25B2064, 
the “Pioneer” and “Leading Goose” R\&D Program of Zhejiang under Grant No. 2025C02110, 
the Public Welfare Research Program of Ningbo under Grant No. 2024S062, 
and the Yongjiang Talent Project of Ningbo under Grant No. 2024A-161-G.

This research was also supported by Meituan.
\end{acks}

\bibliographystyle{ACM-Reference-Format}
\bibliography{sample-base}

@misc{survey_sllms_1,
      title={WavChat: A Survey of Spoken Dialogue Models}, 
      author={Shengpeng Ji and Yifu Chen and Minghui Fang and Jialong Zuo and Jingyu Lu and Hanting Wang and Ziyue Jiang and Long Zhou and Shujie Liu and Xize Cheng and Xiaoda Yang and Zehan Wang and Qian Yang and Jian Li and Yidi Jiang and Jingzhen He and Yunfei Chu and Jin Xu and Zhou Zhao},
      year={2024},
      eprint={2411.13577},
      archivePrefix={arXiv},
      primaryClass={eess.AS},
      url={https://arxiv.org/abs/2411.13577}, 
}

@article{survey_sllms_2,
   title={A Survey on Speech Large Language Models for Understanding},
   volume={20},
   ISSN={1941-0484},
   url={http://dx.doi.org/10.1109/JSTSP.2025.3640535},
   DOI={10.1109/jstsp.2025.3640535},
   number={1},
   journal={IEEE Journal of Selected Topics in Signal Processing},
   publisher={Institute of Electrical and Electronics Engineers (IEEE)},
   author={Peng, Jing and Wang, Yucheng and Li, Bohan and Guo, Yiwei and Wang, Hankun and Fang, YanGui and Xi, Yu and Li, Haoyu and Li, Xu and Zhang, Ke and Wang, Shuai and Yu, Kai},
   year={2026},
   month=jan, pages={2–31} }

@inproceedings{Snapkv,
 author = {Li, Yuhong and Huang, Yingbing and Yang, Bowen and Venkitesh, Bharat and Locatelli, Acyr and Ye, Hanchen and Cai, Tianle and Lewis, Patrick and Chen, Deming},
 booktitle = {Advances in Neural Information Processing Systems},
 doi = {10.52202/079017-0722},
 editor = {A. Globerson and L. Mackey and D. Belgrave and A. Fan and U. Paquet and J. Tomczak and C. Zhang},
 pages = {22947--22970},
 publisher = {Curran Associates, Inc.},
 title = {SnapKV: LLM Knows What You are Looking for Before Generation},
 url = {https://proceedings.neurips.cc/paper_files/paper/2024/file/28ab418242603e0f7323e54185d19bde-Paper-Conference.pdf},
 volume = {37},
 year = {2024}
}

@misc{pyramidKV,
      title={PyramidKV: Dynamic KV Cache Compression based on Pyramidal Information Funneling}, 
      author={Zefan Cai and Yichi Zhang and Bofei Gao and Yuliang Liu and Yucheng Li and Tianyu Liu and Keming Lu and Wayne Xiong and Yue Dong and Junjie Hu and Wen Xiao},
      year={2025},
      eprint={2406.02069},
      archivePrefix={arXiv},
      primaryClass={cs.CL},
      url={https://arxiv.org/abs/2406.02069}, 
}

@misc{ChunkKV,
      title={ChunkKV: Semantic-Preserving KV Cache Compression for Efficient Long-Context LLM Inference}, 
      author={Xiang Liu and Zhenheng Tang and Peijie Dong and Zeyu Li and Yue Liu and Bo Li and Xuming Hu and Xiaowen Chu},
      year={2025},
      eprint={2502.00299},
      archivePrefix={arXiv},
      primaryClass={cs.CL},
      url={https://arxiv.org/abs/2502.00299}, 
}

@inproceedings{
SLLM,
title={Efficient Streaming Language Models with Attention Sinks},
author={Guangxuan Xiao and Yuandong Tian and Beidi Chen and Song Han and Mike Lewis},
booktitle={The Twelfth International Conference on Learning Representations},
year={2024},
url={https://openreview.net/forum?id=NG7sS51zVF}
}

@INPROCEEDINGS{SpeechPrune,
  author={Lin, Yueqian and Fu, Yuzhe and Zhang, Jingyang and Liu, Yudong and Zhang, Jianyi and Sun, Jingwei and Li, Hai Helen and Chen, Yiran},
  booktitle={2025 IEEE International Conference on Multimedia and Expo (ICME)}, 
  title={SpeechPrune: Context-Aware Token Pruning for Speech Information Retrieval}, 
  year={2025},
  volume={},
  number={},
  pages={1-6},
  doi={10.1109/ICME59968.2025.11209113}}

@misc{FastV,
      title={An Image is Worth 1/2 Tokens After Layer 2: Plug-and-Play Inference Acceleration for Large Vision-Language Models}, 
      author={Liang Chen and Haozhe Zhao and Tianyu Liu and Shuai Bai and Junyang Lin and Chang Zhou and Baobao Chang},
      year={2024},
      eprint={2403.06764},
      archivePrefix={arXiv},
      primaryClass={cs.CV},
      url={https://arxiv.org/abs/2403.06764}, 
}

@InProceedings{DyCoke,
    author    = {Tao, Keda and Qin, Can and You, Haoxuan and Sui, Yang and Wang, Huan},
    title     = {DyCoke: Dynamic Compression of Tokens for Fast Video Large Language Models},
    booktitle = {Proceedings of the Computer Vision and Pattern Recognition Conference (CVPR)},
    month     = {June},
    year      = {2025},
    pages     = {18992-19001}
}

@misc{kv_cache_survey_1,
      title={A Survey on Large Language Model Acceleration based on KV Cache Management}, 
      author={Haoyang Li and Yiming Li and Anxin Tian and Tianhao Tang and Zhanchao Xu and Xuejia Chen and Nicole Hu and Wei Dong and Qing Li and Lei Chen},
      year={2025},
      eprint={2412.19442},
      archivePrefix={arXiv},
      primaryClass={cs.AI},
      url={https://arxiv.org/abs/2412.19442}, 
}

@misc{kv_cache_survey_2,
      title={A Survey on Efficient Inference for Large Language Models}, 
      author={Zixuan Zhou and Xuefei Ning and Ke Hong and Tianyu Fu and Jiaming Xu and Shiyao Li and Yuming Lou and Luning Wang and Zhihang Yuan and Xiuhong Li and Shengen Yan and Guohao Dai and Xiao-Ping Zhang and Yuhan Dong and Yu Wang},
      year={2024},
      eprint={2404.14294},
      archivePrefix={arXiv},
      primaryClass={cs.CL},
      url={https://arxiv.org/abs/2404.14294}, 
}

@misc{Whisper,
      title={Robust Speech Recognition via Large-Scale Weak Supervision}, 
      author={Alec Radford and Jong Wook Kim and Tao Xu and Greg Brockman and Christine McLeavey and Ilya Sutskever},
      year={2022},
      eprint={2212.04356},
      archivePrefix={arXiv},
      primaryClass={eess.AS},
      url={https://arxiv.org/abs/2212.04356}, 
}

@article{Qwen3-Omni,
  title={Qwen3-Omni Technical Report},
  author={Jin Xu and Zhifang Guo and Hangrui Hu and Yunfei Chu and Xiong Wang and Jinzheng He and Yuxuan Wang and Xian Shi and Ting He and Xinfa Zhu and Yuanjun Lv and Yongqi Wang and Dake Guo and He Wang and Linhan Ma and Pei Zhang and Xinyu Zhang and Hongkun Hao and Zishan Guo and Baosong Yang and Bin Zhang and Ziyang Ma and Xipin Wei and Shuai Bai and Keqin Chen and Xuejing Liu and Peng Wang and Mingkun Yang and Dayiheng Liu and Xingzhang Ren and Bo Zheng and Rui Men and Fan Zhou and Bowen Yu and Jianxin Yang and Le Yu and Jingren Zhou and Junyang Lin},
  journal={arXiv preprint arXiv:2509.17765},
  year={2025}
}

@misc{Qwen2.5-Omni,
      title={Qwen2.5-Omni Technical Report}, 
      author={Jin Xu and Zhifang Guo and Jinzheng He and Hangrui Hu and Ting He and Shuai Bai and Keqin Chen and Jialin Wang and Yang Fan and Kai Dang and Bin Zhang and Xiong Wang and Yunfei Chu and Junyang Lin},
      year={2025},
      eprint={2503.20215},
      archivePrefix={arXiv},
      primaryClass={cs.CL},
      url={https://arxiv.org/abs/2503.20215}, 
}

@misc{Kimi-Audio,
      title={Kimi-Audio Technical Report}, 
      author={KimiTeam and Ding Ding and Zeqian Ju and Yichong Leng and Songxiang Liu and Tong Liu and Zeyu Shang and Kai Shen and Wei Song and Xu Tan and Heyi Tang and Zhengtao Wang and Chu Wei and Yifei Xin and Xinran Xu and Jianwei Yu and Yutao Zhang and Xinyu Zhou and Y. Charles and Jun Chen and Yanru Chen and Yulun Du and Weiran He and Zhenxing Hu and Guokun Lai and Qingcheng Li and Yangyang Liu and Weidong Sun and Jianzhou Wang and Yuzhi Wang and Yuefeng Wu and Yuxin Wu and Dongchao Yang and Hao Yang and Ying Yang and Zhilin Yang and Aoxiong Yin and Ruibin Yuan and Yutong Zhang and Zaida Zhou},
      year={2025},
      eprint={2504.18425},
      archivePrefix={arXiv},
      primaryClass={eess.AS},
      url={https://arxiv.org/abs/2504.18425}, 
}

@inproceedings{
AudioFlamingo-3,
title={Audio Flamingo 3: Advancing Audio Intelligence with Fully Open Large Audio Language Models},
author={Sreyan Ghosh and Arushi Goel and Jaehyeon Kim and Sonal Kumar and Zhifeng Kong and Sang-gil Lee and Chao-Han Huck Yang and Ramani Duraiswami and Dinesh Manocha and Rafael Valle and Bryan Catanzaro},
booktitle={The Thirty-ninth Annual Conference on Neural Information Processing Systems},
year={2025},
url={https://openreview.net/forum?id=FjByDpDVIO}
}

@misc{AudioFlamingo-2,
      title={Audio Flamingo 2: An Audio-Language Model with Long-Audio Understanding and Expert Reasoning Abilities}, 
      author={Sreyan Ghosh and Zhifeng Kong and Sonal Kumar and S Sakshi and Jaehyeon Kim and Wei Ping and Rafael Valle and Dinesh Manocha and Bryan Catanzaro},
      year={2025},
      eprint={2503.03983},
      archivePrefix={arXiv},
      primaryClass={cs.SD},
      url={https://arxiv.org/abs/2503.03983}, 
}

@misc{KV_cache_expense,
      title={Efficiently Scaling Transformer Inference}, 
      author={Reiner Pope and Sholto Douglas and Aakanksha Chowdhery and Jacob Devlin and James Bradbury and Anselm Levskaya and Jonathan Heek and Kefan Xiao and Shivani Agrawal and Jeff Dean},
      year={2022},
      eprint={2211.05102},
      archivePrefix={arXiv},
      primaryClass={cs.LG},
      url={https://arxiv.org/abs/2211.05102}, 
}

@misc{
Vox-Infinity,
title={Vox-Infinity: Benchmarking the Limits of Long-Context Spoken Language Models},
author={Xize Cheng and Dongjie Fu and Chenyuhao Wen and Tao Jin and Hai Yu and Di Cao and Qinying Liu and Yexin Yang and Zehan Wang and Shengpeng Ji and Siqi Zheng and Xu Tan and Zhou Zhao},
year={2026},
url={https://openreview.net/forum?id=6dKwqnT7bu}
}

@inproceedings{MMAU,
 author = {Sakshi, Sakshi and Tyagi, Utkarsh and Kumar, Sonal and Seth, Ashish and Selvakumar, Ramaneswaran and Nieto, Oriol and Duraiswami, Ramani and Ghosh, Sreyan and Manocha, Dinesh},
 booktitle = {International Conference on Learning Representations},
 editor = {Y. Yue and A. Garg and N. Peng and F. Sha and R. Yu},
 pages = {84929--84964},
 title = {MMAU: A Massive Multi-Task Audio Understanding and Reasoning Benchmark},
 url = {https://proceedings.iclr.cc/paper_files/paper/2025/file/d36f208919582785db965fe648b9fe59-Paper-Conference.pdf},
 volume = {2025},
 year = {2025}
}

@inproceedings{
MMSU,
title={{MMSU}: A Massive Multi-task Spoken Language Understanding and Reasoning Benchmark},
author={Dingdong WANG and Jincenzi Wu and Junan Li and Dongchao Yang and Xueyuan Chen and Tianhua Zhang and Helen M. Meng},
booktitle={The Fourteenth International Conference on Learning Representations},
year={2026},
url={https://openreview.net/forum?id=yHzCDP1tXw}
}

@inproceedings{
MMAR,
title={{MMAR}: A Challenging Benchmark for Deep Reasoning in Speech, Audio, Music, and Their Mix},
author={Ziyang Ma and Yinghao Ma and Yanqiao Zhu and Chen Yang and Yi-Wen Chao and Ruiyang Xu and Wenxi Chen and Yuanzhe Chen and Zhuo Chen and Jian Cong and Kai Li and Keliang Li and Siyou Li and Xinfeng Li and Xiquan Li and Zheng Lian and Yuzhe Liang and Minghao Liu and Zhikang Niu and Tianrui Wang and Yuping Wang and Yuxuan Wang and Yihao Wu and Guanrou Yang and Jianwei Yu and Ruibin Yuan and Zhisheng Zheng and Ziya Zhou and Haina Zhu and Wei Xue and Emmanouil Benetos and Kai Yu and EngSiong Chng and Xie Chen},
booktitle={The Thirty-ninth Annual Conference on Neural Information Processing Systems Datasets and Benchmarks Track},
year={2025},
url={https://openreview.net/forum?id=fgmrBJemlQ}
}

@misc{AudioMarathon,
      title={AudioMarathon: A Comprehensive Benchmark for Long-Context Audio Understanding and Efficiency in Audio LLMs}, 
      author={Peize He and Zichen Wen and Yubo Wang and Yuxuan Wang and Xiaoqian Liu and Jiajie Huang and Zehui Lei and Zhuangcheng Gu and Xiangqi Jin and Jiabing Yang and Kai Li and Zhifei Liu and Weijia Li and Cunxiang Wang and Conghui He and Linfeng Zhang},
      year={2025},
      eprint={2510.07293},
      archivePrefix={arXiv},
      primaryClass={cs.SD},
      url={https://arxiv.org/abs/2510.07293}, 
}

@inproceedings{LOOK_M,
    title = "{LOOK}-{M}: Look-Once Optimization in {KV} Cache for Efficient Multimodal Long-Context Inference",
    author = "Wan, Zhongwei  and
      Wu, Ziang  and
      Liu, Che  and
      Huang, Jinfa  and
      Zhu, Zhihong  and
      Jin, Peng  and
      Wang, Longyue  and
      Yuan, Li",
    editor = "Al-Onaizan, Yaser  and
      Bansal, Mohit  and
      Chen, Yun-Nung",
    booktitle = "Findings of the Association for Computational Linguistics: EMNLP 2024",
    month = nov,
    year = "2024",
    address = "Miami, Florida, USA",
    publisher = "Association for Computational Linguistics",
    url = "https://aclanthology.org/2024.findings-emnlp.235/",
    doi = "10.18653/v1/2024.findings-emnlp.235",
    pages = "4065--4078",
}

@inproceedings{
H2O,
title={H2O: Heavy-Hitter Oracle for Efficient Generative Inference of Large Language Models},
author={Zhenyu Zhang and Ying Sheng and Tianyi Zhou and Tianlong Chen and Lianmin Zheng and Ruisi Cai and Zhao Song and Yuandong Tian and Christopher Re and Clark Barrett and Zhangyang Wang and Beidi Chen},
booktitle={Thirty-seventh Conference on Neural Information Processing Systems},
year={2023},
url={https://openreview.net/forum?id=RkRrPp7GKO}
}

@misc{GLM-4-Voice,
      title={GLM-4-Voice: Towards Intelligent and Human-Like End-to-End Spoken Chatbot}, 
      author={Aohan Zeng and Zhengxiao Du and Mingdao Liu and Kedong Wang and Shengmin Jiang and Lei Zhao and Yuxiao Dong and Jie Tang},
      year={2024},
      eprint={2412.02612},
      archivePrefix={arXiv},
      primaryClass={cs.CL},
      url={https://arxiv.org/abs/2412.02612}, 
}

@misc{prompt_compression_survey_1,
      title={Prompt Compression for Large Language Models: A Survey}, 
      author={Zongqian Li and Yinhong Liu and Yixuan Su and Nigel Collier},
      year={2024},
      eprint={2410.12388},
      archivePrefix={arXiv},
      primaryClass={cs.CL},
      url={https://arxiv.org/abs/2410.12388}, 
}

@inproceedings{sllm_survey_3,
    title = "Recent Advances in Speech Language Models: A Survey",
    author = "Cui, Wenqian  and
      Yu, Dianzhi  and
      Jiao, Xiaoqi  and
      Meng, Ziqiao  and
      Zhang, Guangyan  and
      Wang, Qichao  and
      Guo, Steven Y.  and
      King, Irwin",
    editor = "Che, Wanxiang  and
      Nabende, Joyce  and
      Shutova, Ekaterina  and
      Pilehvar, Mohammad Taher",
    booktitle = "Proceedings of the 63rd Annual Meeting of the Association for Computational Linguistics (Volume 1: Long Papers)",
    month = jul,
    year = "2025",
    address = "Vienna, Austria",
    publisher = "Association for Computational Linguistics",
    url = "https://aclanthology.org/2025.acl-long.682/",
    doi = "10.18653/v1/2025.acl-long.682",
    pages = "13943--13970",
    ISBN = "979-8-89176-251-0"
}

@misc{BLAB,
      title={BLAB: Brutally Long Audio Bench}, 
      author={Orevaoghene Ahia and Martijn Bartelds and Kabir Ahuja and Hila Gonen and Valentin Hofmann and Siddhant Arora and Shuyue Stella Li and Vishal Puttagunta and Mofetoluwa Adeyemi and Charishma Buchireddy and Ben Walls and Noah Bennett and Shinji Watanabe and Noah A. Smith and Yulia Tsvetkov and Sachin Kumar},
      year={2025},
      eprint={2505.03054},
      archivePrefix={arXiv},
      primaryClass={cs.AI},
      url={https://arxiv.org/abs/2505.03054}, 
}

@misc{kv_cache_review_1,
      title={KV Cache Compression for Inference Efficiency in LLMs: A Review}, 
      author={Yanyu Liu and Jingying Fu and Sixiang Liu and Yitian Zou and You Fu and Jiehan Zhou and Shouhua Zhang},
      year={2025},
      eprint={2508.06297},
      archivePrefix={arXiv},
      primaryClass={cs.DC},
      url={https://arxiv.org/abs/2508.06297}, 
}

@misc{d2o,
      title={D2O: Dynamic Discriminative Operations for Efficient Long-Context Inference of Large Language Models}, 
      author={Zhongwei Wan and Xinjian Wu and Yu Zhang and Yi Xin and Chaofan Tao and Zhihong Zhu and Xin Wang and Siqi Luo and Jing Xiong and Longyue Wang and Mi Zhang},
      year={2025},
      eprint={2406.13035},
      archivePrefix={arXiv},
      primaryClass={cs.CL},
      url={https://arxiv.org/abs/2406.13035}, 
}

@misc{fu2026characterspeechleveragingroleplaying,
      title={Character Beyond Speech: Leveraging Role-Playing Evaluation in Audio Large Language Models via Reinforcement Learning},
      author={Dongjie Fu and Fangming Feng and Xize Cheng and Linjun Li and Zhou Zhao and Tao Jin},
      year={2026},
      eprint={2604.13804},
      archivePrefix={arXiv},
      primaryClass={cs.LG},
      url={https://arxiv.org/abs/2604.13804},
}

@inproceedings{fu-etal-2025-pachat,
    title = {{PACHAT}: Persona-Aware Speech Assistant for Multi-party Dialogue},
    author = {Fu, Dongjie and Cheng, Xize and Li, Linjun and Yang, Xiaoda and Yang, Lujia and Jin, Tao},
    booktitle = {Proceedings of the 2025 Conference on Empirical Methods in Natural Language Processing},
    month = nov,
    year = {2025},
    address = {Suzhou, China},
    publisher = {Association for Computational Linguistics},
    url = {https://aclanthology.org/2025.emnlp-main.1492/},
    doi = {10.18653/v1/2025.emnlp-main.1492},
    pages = {29325--29342},
}

@misc{cao2026xopdcrossmodalonpolicydistillation,
      title={X-OPD: Cross-Modal On-Policy Distillation for Capability Alignment in Speech LLMs},
      author={Di Cao and Dongjie Fu and Hai Yu and Siqi Zheng and Xu Tan and Tao Jin},
      year={2026},
      eprint={2603.24596},
      archivePrefix={arXiv},
      primaryClass={eess.AS},
      url={https://arxiv.org/abs/2603.24596},
}

\end{document}